\documentclass[conference]{IEEEtran}
\IEEEoverridecommandlockouts

\usepackage{cite}
\usepackage{amsmath,amssymb,amsfonts}
\usepackage{graphicx}
\usepackage{textcomp}
\usepackage{xcolor}
\usepackage{booktabs}
\usepackage{array}
\usepackage{multirow}
\usepackage{dblfloatfix}

\begin{document}

\title{Does an Illumination Prior Help Face-Swap Detection?\\
A Controlled Study of Temporal Self-Blended Images}

\author{%
\IEEEauthorblockN{Danil Davydov\IEEEauthorrefmark{1},
Bader Rasheed\IEEEauthorrefmark{2},
Dmitriy Vatolin\IEEEauthorrefmark{2}}
\IEEEauthorblockA{%
\IEEEauthorrefmark{1}\textit{Innopolis University}, Innopolis, Russia\\
\IEEEauthorrefmark{2}\textit{Laboratory of Innovative Technologies for Processing
Video Content, Innopolis University}, Innopolis, Russia\\
\texttt{\{d.davydov, b.rasheed, d.vatolin\}@innopolis.university}}}

\maketitle

\begin{abstract}
Self-blending is the dominant training-data prior in face-swap detection, but it
supervises only the blending boundary. We ask whether adding a \emph{photometric}
prior at the same locus helps: Temporal Self-Blended Images (T-SBI) transfer the
illumination statistics of one frame onto a self-blended copy of another frame from
the same video, gated on a controlled luminance delta $\Delta L$ in LAB space, so that
illumination inconsistency becomes a learnable training signal. We evaluate T-SBI as a
controlled probe rather than as a proposed improvement. A five-regime ablation
factorizes supervised, self-blended, and temporally self-blended training; a
three-seed $\Delta L$-quartile experiment isolates the illumination delta itself by
training on the upper versus lower quartile of accepted pairs. The result is a
consistent null on the targeted mechanism: frame-level AUC differences between
high-$\Delta L$ and low-$\Delta L$ training lie within the combined seed standard
deviation on all four evaluation corpora, and a failure taxonomy over
$506{,}328$ attribute-binned rows finds \emph{no} preferential failure-rate reduction
on harsh-illumination bins. What T-SBI does do is shift the score distribution: the
optimal decision threshold moves down by ${\approx}0.34$ on FF++ and ${\approx}0.30$ on
Celeb-DF, which reverses the apparent verdict when models are compared at the
conventional threshold of $0.5$. The one robust benefit lies elsewhere than intended:
under heavy JPEG compression the T-SBI variant retains frame AUC $0.780$ on DFDC against
$0.696$ for its counterpart, consistent with a shift toward low-frequency cues rather
than illumination-specific robustness. We report the operating-point analysis that
resolves the threshold contradiction, show that F1 is degenerate under cross-dataset
imbalance, and argue that photometric training priors must be validated against the
attribute they target rather than against aggregate accuracy.
\end{abstract}


\section{Introduction}
Face-swap detectors that learn generator-specific texture artefacts transfer poorly to
unseen manipulations~\cite{rossler2019faceforensics,nadimpalli2022crossdataset}. The
most effective known remedy is a training-data prior rather than an architectural one:
Face X-Ray~\cite{li2020facexray} and Self-Blended Images (SBI)~\cite{shiohara2022sbi}
synthesize fakes by blending an image with a perturbed copy of itself, forcing the
classifier onto the blending boundary---a cue common to all compositing pipelines---and
achieving strong cross-dataset transfer without labelled fakes.

The blending boundary is not, however, the only physical trace a face swap leaves. The
photometric-forensics literature holds that compositing two faces recorded under
different illumination produces inconsistencies that physical scene constraints would
forbid~\cite{johnson2007lightingforensics,kee2013shadows}. This suggests an obvious
extension: build the self-blended training pair from \emph{two frames of the same
video} chosen to differ in illumination, so that the synthetic fake carries a
photometric inconsistency in addition to a blending boundary. We call this Temporal
Self-Blended Images (T-SBI).

The contribution of this paper is not the generator but the \emph{measurement}. It is
easy to add such a prior and report an aggregate accuracy delta; it is much harder to
show that any observed change is caused by the mechanism the prior was designed around.
We therefore treat T-SBI as a controlled probe and ask three separable questions: does
the temporal-illumination component change detection performance at all; is the
illumination delta of the training pairs the active ingredient; and does the model
become preferentially more robust on the image attribute---illumination---that the
prior targets? Our contributions are:

\begin{itemize}
\item A T-SBI generator with explicit pair-level illumination gating in LAB space, and
a seven-variant shortcut ablation verifying that no class-asymmetric pipeline artefact
confounds the comparison against classic SBI.
\item A three-seed $\Delta L$-quartile experiment isolating the training-pair
illumination delta. High-$\Delta L$ and low-$\Delta L$ training differ by less than the
combined seed standard deviation on every corpus; a video-level direction reversal on
DFDC is shown to be a calibration-mediated aggregation artefact.
\item A failure taxonomy over $506{,}328$ attribute-binned rows (nine models, four
datasets, nine attribute axes) showing that T-SBI produces no preferential
failure-rate reduction on harsh-illumination bins---the targeted mechanism is
unsupported---and an operating-point analysis identifying a ${\approx}0.30$ downward
shift in optimal threshold, plus a compression-robustness gain of $+0.084$ AUC on DFDC,
as the effects that are actually present.
\end{itemize}

\section{Related Work}
\textbf{Blending priors.} Face X-Ray~\cite{li2020facexray} localizes forgeries by
detecting blending boundaries; SBI~\cite{shiohara2022sbi} removes the dependence on
labelled fakes by self-blending a single source image and remains the reference
cross-dataset baseline, which we adopt.

\textbf{Photometric forensics.} Illumination direction and shadow geometry constrain
plausible image formation, and mismatches indicate
manipulation~\cite{johnson2007lightingforensics,kee2013shadows}; automated
lighting-inconsistency detectors follow this line~\cite{wu2022lighting}. These methods
are typically applied at inference. T-SBI instead injects the photometric constraint at
training time, which is where the blending prior already lives.

\textbf{Evaluation under imbalance.} Cross-dataset benchmarks are heavily imbalanced, and
threshold-$0.5$ metrics conflate calibration with
capability~\cite{nadimpalli2022crossdataset,guo2017calibration}; we therefore report
imbalance-robust threshold sweeps~\cite{matthews1975mcc}, which prove decisive here.

\section{Method}
\subsection{Temporal self-blended image generation}
Each T-SBI pair is built from two frames of the same source video at distinct time
positions. MediaPipe FaceMesh~\cite{lugaresi2019mediapipe} landmarks define a feathered
convex-hull mask over the target face (fallback: axis-aligned ellipse). The source
frame's illumination statistics---L-channel mean, L-channel standard deviation, and a
low-frequency luminance term inside the mask---are transferred onto a self-blended copy
of the target frame, preserving the target's geometry, identity, expression, and pose so
that the training signal is \emph{differential illumination} rather than incidental
geometric inconsistency. We use luminance-only histogram matching with percentile trimming and
envelope-based fitting, which suppresses three observed failure modes (flat unicolour
faces, blowouts, band artefacts). Pairs are rejected if the source and target boxes
differ by ${>}1.3\times$ in scale or ${>}35\%$ in centre offset.

\subsection{Pair-level illumination gating}
Because the hypothesis is that the illumination delta is the active ingredient, each
candidate pair is gated explicitly. Let $\mathcal{M}$ denote the pixels inside the face
box and $L_I(p)$ the CIELAB lightness of frame $I$ at pixel $p$, with within-box moments
\begin{equation}
\begin{aligned}
\mu_L(I) &= \frac{1}{|\mathcal{M}|}\sum_{p\in\mathcal{M}} L_I(p), \\[2pt]
\sigma_L^{2}(I) &= \frac{1}{|\mathcal{M}|}\sum_{p\in\mathcal{M}}
\bigl(L_I(p)-\mu_L(I)\bigr)^{2}.
\end{aligned}
\label{eq:moments}
\end{equation}
The pair-level deltas between source frame $I_s$ and target frame $I_t$ are
\begin{equation}
\begin{aligned}
\Delta L_{\text{mean}}=\bigl|\mu_L(I_s)-\mu_L(I_t)\bigr|,
\;\; \\[2pt]
\Delta L_{\text{std}}=\bigl|\sigma_L(I_s)-\sigma_L(I_t)\bigr| ,
\label{eq:deltas}
\end{aligned}
\end{equation}
so $\Delta L_{\text{std}}$ is the difference of the two per-frame standard
deviations---a change in how spread out the shading is---not the standard deviation of a
difference image. Statistics are taken over the detector's face box, and lightness is on
the $8$-bit OpenCV LAB scale $L\in[0,255]$, so the thresholds below correspond to roughly
$2.4$ and $1.2$ units of CIE $L^{*}$. A pair is accepted if
\begin{equation}
\Delta L_{\text{mean}} \ge 6 \quad\text{or}\quad \Delta L_{\text{std}} \ge 3 .
\label{eq:gate}
\end{equation}
The dual criterion covers a global luminance offset and a regional illumination
redistribution respectively. To avoid silently discarding evenly-lit videos, a relax
fallback accepts the most-different available pair when no pair from a source video
clears Eq.~\eqref{eq:gate}. Every row records $\Delta L_{\text{mean}}$,
$\Delta L_{\text{std}}$, and a relax flag, enabling the quartile decomposition below.

The realized pool is itself informative: across $7{,}320$ accepted pairs the median
$\Delta L_{\text{mean}}$ is $5.62$---\emph{below} the design threshold---and $22.3\%$ of
pairs entered only via the relax fallback. This is not a property of the source
material: over the $732$ contributing videos the median per-video maximum
$\Delta L_{\text{mean}}$ is $10.13$ and $89\%$ contain at least one pair clearing the
gate. The sampler, not the corpus, is the binding constraint.

\subsection{Five-regime ablation}
Five training regimes factorize the supervised, self-blended, and temporally
self-blended components: \textbf{A} (real + FF++ supervised fakes); \textbf{B\_pure}
(real + classic SBI fakes); \textbf{B\_mix} (A + SBI); \textbf{C\_pure} (real + SBI +
T-SBI); \textbf{C\_mix} (A + SBI + T-SBI). The clean contrast for the
temporal-illumination signal is \textbf{B\_mix vs C\_mix} (and B\_pure vs C\_pure):
identical recipe, hyperparameters, real-class data, and SBI fakes, differing only by the
addition of T-SBI pairs. All regimes share an EfficientNet-B4
backbone~\cite{tan2019efficientnet}, AdamW with cosine warmup, $380\times380$ input,
twenty epochs, and class-balanced batch sampling. Per-image JPEG quality is randomized
over $[75,98]$ for \emph{all} real and synthetic samples in all regimes, eliminating
compression history as a class-asymmetric shortcut.

\subsection{Shortcut ablation}
To verify that any B-vs-C difference reflects training data rather than a pipeline
artefact, the T-SBI pipeline is decomposed into seven null variants (N0--N6) in which the
active step is replaced by an identity operation, with a parallel suite (P0--P5) for
classic SBI. A one-epoch classifier is trained on (real, null-output) and separability
measured two-sided as $|\mathrm{AUC}-0.5|$; ${\le}0.05$ indicates no class-asymmetric
signal. The photometric-transfer variants driving the headline contrast (N3--N6) all fall
inside the band. Two sit outside: N0 (JPEG round-trip, $0.428$), present identically in
both B and C regimes and therefore cancelling in the contrast, and N2 (adjacent-frame
pairing, $0.431$), unique to T-SBI and reported as a caveat.

\subsection{Decisive diagnostic: the $\Delta L$-quartile experiment}
The ablation above changes \emph{whether} T-SBI pairs are present; it does not isolate
the illumination delta. We therefore split the accepted pool at its quartiles and build
two C\_mix-recipe training sets that are byte-identical except for the T-SBI portion:
\textbf{HIGHDL} (upper quartile, $1{,}830$ pairs, mean $\Delta L_{\text{mean}}=12.43$)
and \textbf{LOWDL} (lower quartile, $1{,}830$ pairs, mean $1.57$). Real frames,
supervised fakes, and classic SBI fakes ($24{,}676$ rows) are held identical. Each
variant is trained for twenty epochs across three seeds ($42$, $123$, $235$) and
evaluated on FF++ (in-distribution) and Celeb-DF~\cite{li2020celebdf},
DFDC~\cite{dolhansky2020dfdc}, and DFF~\cite{song2023dff} (cross-dataset). If the
illumination delta is the active ingredient, HIGHDL should separate from LOWDL by more
than seed noise.

\subsection{Failure taxonomy}
For every crop in every evaluation set we compute nine image attributes (blur,
illumination flatness, illumination harshness, yaw, pitch, eye state, gaze deviation,
crop tightness, frame-edge contact), binned by within-dataset quantiles, yielding
$506{,}328$ rows over nine models and four datasets. For each (model, dataset, axis) we
build a contingency table of failure status against attribute bin, apply Pearson's
$\chi^2$, and report Cram\'er's $V$ as the effect size---the meaningful summary, since
at these sample sizes negligible effects are still highly significant. We adopt Cohen's
conventions ($V<0.1$ negligible)~\cite{cohen1988}.

\section{Results}
\subsection{Five-regime ablation: no consistent T-SBI direction}
Table~\ref{tab:regimes} reports cross-dataset AUC and calibration. Three observations
follow. First, mixed regimes beat the supervised baseline on DFDC by ${\approx}+0.04$
while A retains the Celeb-DF AUC lead---the canonical SBI-family profile. Second, pure
self-blend regimes collapse by $0.13$--$0.24$ AUC, and their in-distribution FF++ AUC
($0.771$/$0.732$) with ECE ${\approx}0.43$--$0.48$ shows the failure is \emph{global},
not cross-distributional: mixed supervision is necessary for these recipes to produce a
deployable detector at all. Third, and central here, the C-vs-B contrast that isolates
T-SBI has no consistent direction: C\_mix exceeds B\_mix by $+0.014$ on Celeb-DF but
trails by $-0.003$ on DFDC, and C\_pure trails B\_pure on both. The ablation licenses no
universal claim about T-SBI.

Calibration tells a cleaner story than AUC: baseline A is severely miscalibrated despite
its Celeb-DF AUC lead (frame ECE $0.382$ vs $0.030$ for C\_mix, for $0.044$ AUC).

\begin{table}[t]
\caption{Five-regime cross-dataset AUC (video-level mean aggregation; DFF frame-level)
and frame-level ECE. The C-vs-B contrast isolating T-SBI has no consistent sign.
Single-seed; see Limitations.}
\label{tab:regimes}
\centering
\footnotesize
\setlength{\tabcolsep}{4pt}
\begin{tabular}{lrrrrrr}
\toprule
& \multicolumn{4}{c}{\textbf{AUC}} & \multicolumn{2}{c}{\textbf{ECE}} \\
\cmidrule(lr){2-5}\cmidrule(lr){6-7}
\textbf{Regime} & \textbf{FF++} & \textbf{C-DF} & \textbf{DFDC} & \textbf{DFF}
& \textbf{C-DF} & \textbf{DFDC} \\
\midrule
A (supervised)   & $0.991$ & $\mathbf{0.883}$ & $0.824$ & $0.540$ & $0.382$ & $0.482$ \\
B\_pure          & $0.771$ & $0.622$ & $0.691$ & $\mathbf{0.566}$ & $0.072$ & $0.298$ \\
B\_mix           & $\mathbf{0.993}$ & $0.825$ & $\mathbf{0.869}$ & $\mathbf{0.566}$ & $0.041$ & $\mathbf{0.281}$ \\
C\_pure          & $0.732$ & $0.605$ & $0.625$ & $0.546$ & $0.080$ & $0.300$ \\
C\_mix           & $0.990$ & $0.839$ & $0.866$ & $0.543$ & $\mathbf{0.030}$ & $0.305$ \\
\midrule
\textit{SBI (ref.)} & $0.879$ & $0.907$ & $0.884$ & $0.614$ & --- & --- \\
\bottomrule
\end{tabular}
\end{table}

All regimes sit near chance on DFF ($0.54$--$0.57$): SBI and T-SBI supervise blending
boundaries and illumination differences across them, neither of which exists in fully
generated diffusion faces, so our claims are scoped to the face-swap family.

\subsection{The illumination delta is not the active ingredient}
Table~\ref{tab:dl} reports the three-seed $\Delta L$-quartile result. Every frame-level
difference lies inside the combined seed standard deviation: the ratio
$|\Delta|/\mathrm{std}_{\text{comb}}$ is $0.71$ (Celeb-DF), $0.54$ (DFDC), $0.08$ (DFF),
and $0.12$ (FF++), none approaching a defensible threshold of ${\approx}2$. The nominal
sign favours HIGHDL on the two face-swap OOD corpora and is essentially zero elsewhere,
but no cell supports a claim of difference. Training on pairs with an order-of-magnitude
larger illumination delta ($12.43$ vs $1.57$) does not measurably change frame-level
detection.

\begin{table}[t]
\caption{Three-seed $\Delta L$-quartile diagnostic, frame-level AUC (mean $\pm$ sample
std over seeds $42/123/235$). All differences lie within the combined seed standard
deviation.}
\label{tab:dl}
\centering
\footnotesize
\setlength{\tabcolsep}{5pt}
\begin{tabular}{lrrrr}
\toprule
\textbf{Dataset} & \textbf{HIGHDL} & \textbf{LOWDL} & \textbf{$\Delta$} & \textbf{$|\Delta|/\sigma$} \\
\midrule
Celeb-DF & $0.675 \pm 0.017$ & $0.652 \pm 0.027$ & $+0.022$ & $0.71$ \\
DFDC     & $0.642 \pm 0.004$ & $0.637 \pm 0.008$ & $+0.005$ & $0.54$ \\
DFF      & $0.553 \pm 0.005$ & $0.554 \pm 0.012$ & $-0.001$ & $0.08$ \\
FF++     & $0.784 \pm 0.046$ & $0.790 \pm 0.015$ & $-0.006$ & $0.12$ \\
\bottomrule
\end{tabular}
\end{table}

\begin{figure*}[!tb]
\centering
\includegraphics[width=\textwidth]{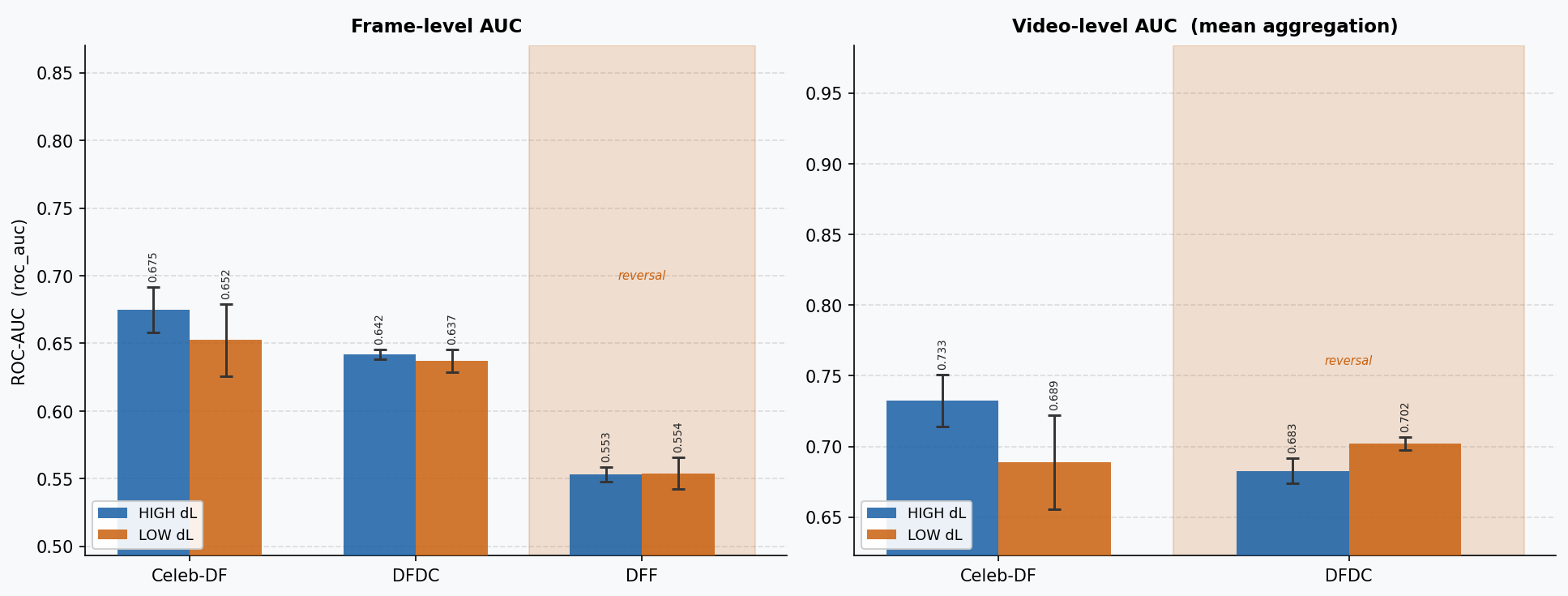}
\caption{HIGHDL vs LOWDL AUC with three-seed error bars. \emph{Left:} frame level,
four datasets---HIGHDL is at or above LOWDL on every OOD corpus, but all margins are
small relative to seed variance and both variants discriminate only weakly
($0.55$--$0.67$). \emph{Right:} video level (mean aggregation), three datasets---the DFDC
comparison reverses, coinciding with severe frame-level miscalibration on that corpus
(ECE ${\approx}0.42$ for both).}
\label{fig:dl}
\end{figure*}

Video-level aggregation appears to contradict this: HIGHDL wins all three strategies
(max/mean/vote) on Celeb-DF, LOWDL wins all three on DFDC, with the DFDC mean and vote
cells reaching ratios of $1.92$ and $2.09$---the study's strongest signal, pointing
\emph{against} HIGHDL. The mechanism is calibration, not discrimination. Both variants are
severely miscalibrated on DFDC (frame ECE $0.419$ and $0.430$, with large seed variance),
and mean and vote aggregation depend on the per-frame score \emph{distribution}, so
calibration differences compound across frames. Max---using only the top-scoring frame,
the closest to a rank-based aggregator---shrinks the gap to $0.60$, matching the
frame-level near-tie. The reversal is thus a miscalibration-mediated aggregation artefact,
not evidence that low-$\Delta L$ training yields better features (Fig.~\ref{fig:dl}).
Table~\ref{tab:dlvid} gives the video-level numbers behind this argument.

\begin{table}[t]
\caption{Three-seed $\Delta L$-quartile diagnostic, video-level AUC by aggregation
strategy. Celeb-DF favours HIGHDL under all three strategies and DFDC favours LOWDL;
the DFDC gap is largest for the two distribution-dependent aggregators (mean, vote) and
collapses for max, which is the closest to a rank-based aggregator.}
\label{tab:dlvid}
\centering
\footnotesize
\setlength{\tabcolsep}{4pt}
\begin{tabular}{llrrrr}
\toprule
\textbf{Dataset} & \textbf{Aggr.} & \textbf{HIGHDL} & \textbf{LOWDL} & \textbf{$\Delta$} & \textbf{$|\Delta|/\sigma$} \\
\midrule
\multirow{3}{*}{Celeb-DF} & max  & $0.727 \pm 0.087$ & $0.708 \pm 0.087$ & $+0.019$ & $0.15$ \\
                          & mean & $0.733 \pm 0.018$ & $0.689 \pm 0.033$ & $+0.044$ & $1.15$ \\
                          & vote & $0.661 \pm 0.061$ & $0.633 \pm 0.031$ & $+0.028$ & $0.41$ \\
\midrule
\multirow{3}{*}{DFDC}     & max  & $0.668 \pm 0.014$ & $0.691 \pm 0.036$ & $-0.023$ & $0.60$ \\
                          & mean & $0.683 \pm 0.009$ & $0.702 \pm 0.005$ & $-0.019$ & $1.92$ \\
                          & vote & $0.667 \pm 0.009$ & $0.704 \pm 0.015$ & $-0.037$ & $2.09$ \\
\midrule
\multirow{3}{*}{FF++}     & max  & $0.918 \pm 0.015$ & $0.917 \pm 0.021$ & $+0.001$ & $0.05$ \\
                          & mean & $0.903 \pm 0.024$ & $0.898 \pm 0.019$ & $+0.005$ & $0.17$ \\
                          & vote & $0.903 \pm 0.022$ & $0.897 \pm 0.032$ & $+0.006$ & $0.16$ \\
\bottomrule
\end{tabular}
\end{table}

\subsection{A dataset-level illumination control}
A natural alternative explanation for the Celeb-DF/DFDC asymmetry is that Celeb-DF is
illumination-static, leaving a within-clip illumination feature nothing to act on.
Measured on real crops this is false: Celeb-DF has the \emph{largest} median within-video
$\Delta L_{\text{mean}}$ p90 ($9.97$, vs $8.91$ FF++ and $9.76$ DFDC) and the smallest
static-clip fraction ($0.15$, vs $0.29$ and $0.21$). DFDC differs instead on orthogonal
axes: shadow-structure variation ($\Delta L_{\text{std}}$ p90 $=6.5$ vs ${\approx}4.2$)
and between-video diversity (p95--p05 of per-video mean $L$: $119.4$ vs $76$--$82$). The
variable T-SBI targets does not separate these corpora, ruling out within-video
illumination distribution as the mediator.

\subsection{The mechanism is a threshold shift, not illumination robustness}
Comparing B\_mix and C\_mix at threshold $0.5$ suggests T-SBI \emph{hurts}: failure rate
rises from $12.8\%$ to $15.9\%$ (Celeb-DF), $35.1\%$ to $42.3\%$ (DFDC), $28.0\%$ to
$34.3\%$ (DFF). Bootstrapped AUC says the opposite on Celeb-DF, with non-overlapping
$95\%$ CIs ($0.736$--$0.742$ vs $0.751$--$0.756$). A full threshold sweep resolves the
contradiction, and the resolution is the paper's methodological point.

First, F1 is unusable here. At $86\%/14\%$ (Celeb-DF), $80\%/20\%$ (DFDC), and
$72\%/28\%$ (DFF) imbalance, a degenerate all-real classifier scores F1 ${\approx}0.92$
on Celeb-DF; empirically, argmax-F1 thresholds collapsed to the grid floor ($0.01$) for
$23$ of $31$ (model, dataset) pairs. We therefore use balanced accuracy, MCC, and
Youden's J over $99$ thresholds in $[0.01,0.99]$.

\begin{figure}[t]
\centering
\includegraphics[width=\columnwidth]{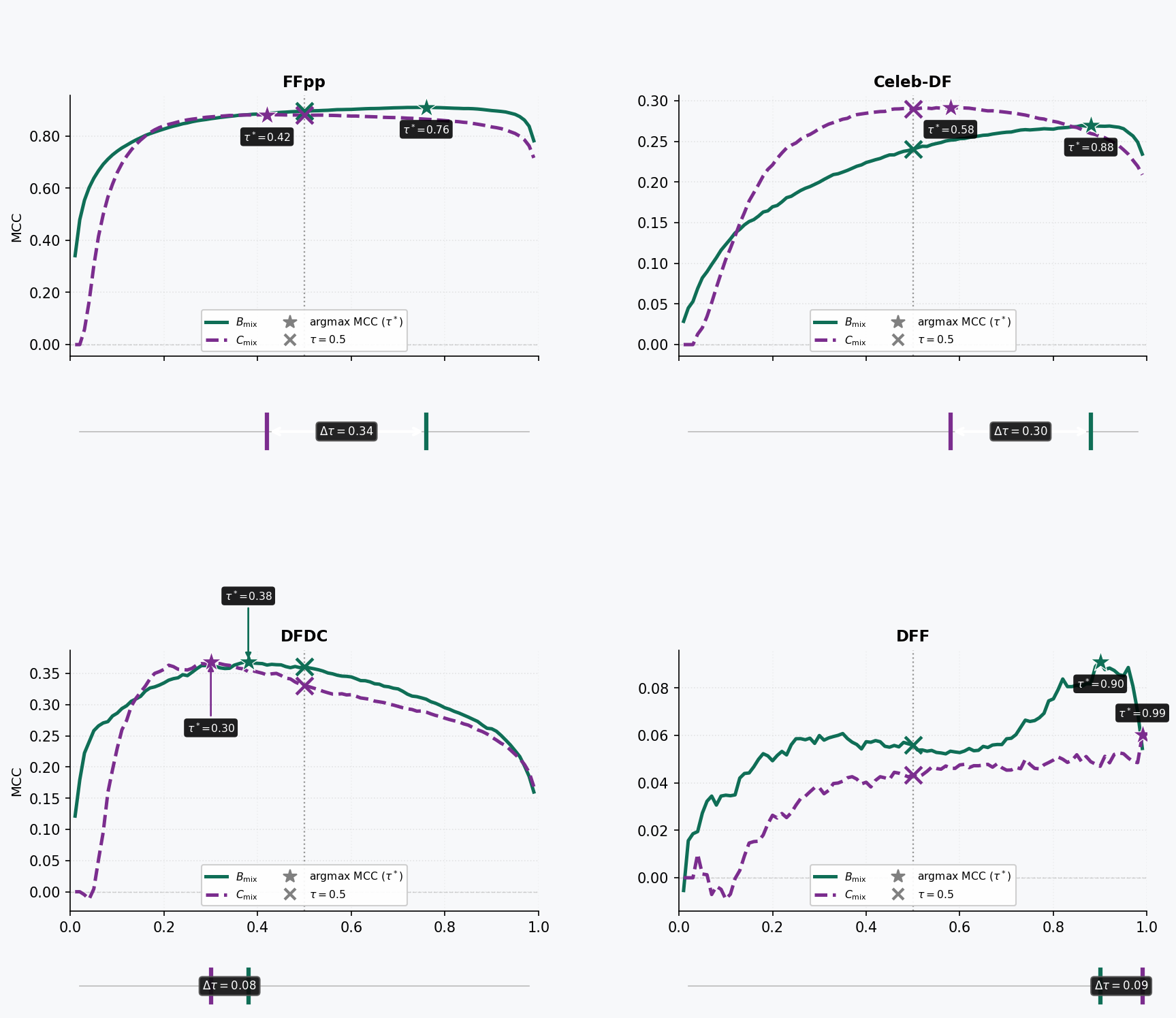}
\caption{MCC versus decision threshold for B\_mix and C\_mix, per dataset. Stars mark
each model's argmax; crosses mark threshold $0.5$. The T-SBI variant's optimum sits
${\approx}0.30$--$0.34$ lower on FF++ and Celeb-DF, which is what makes threshold-$0.5$
comparisons misleading.}
\label{fig:sweep}
\end{figure}

Second, T-SBI moves the operating point. B\_mix's argmax-MCC threshold on Celeb-DF is
$0.88$ while C\_mix's is $0.58$---a shift of ${\approx}0.30$; on FF++ the shift is
$0.76 \to 0.42$, i.e.\ ${\approx}0.34$ (Table~\ref{tab:thr}, Fig.~\ref{fig:sweep}). The
shift attenuates to ${\approx}0.08$--$0.09$ with inconsistent sign on DFDC and DFF.
Where the shift is large, evaluating at $0.5$ systematically penalizes B\_mix (whose
optimum is far above $0.5$) and flatters C\_mix (whose optimum sits near it). The
confusion decomposition makes this concrete: on Celeb-DF at $0.5$, B\_mix has a $0.9\%$
false-positive and $80\%$ false-negative rate, C\_mix $7.9\%$ and $58\%$. T-SBI trades
real-precision for fake-recall; it does not uniformly degrade or improve capability.

\begin{table}[t]
\caption{Threshold-sweep optima (frame level). ``@thr'' is the threshold attaining the
optimum. T-SBI (C\_mix) lowers the optimal operating point substantially on FF++ and
Celeb-DF, marginally elsewhere.}
\label{tab:thr}
\centering
\footnotesize
\setlength{\tabcolsep}{4pt}
\begin{tabular}{llrrrr}
\toprule
\textbf{Dataset} & \textbf{Model} & \textbf{bal-acc} & \textbf{MCC} & \textbf{@thr} & \textbf{J} \\
\midrule
\multirow{2}{*}{FF++}     & B\_mix & $0.968$ & $0.910$ & $0.76$ & $0.935$ \\
                          & C\_mix & $0.961$ & $0.881$ & $0.42$ & $0.922$ \\
\midrule
\multirow{2}{*}{Celeb-DF} & B\_mix & $0.664$ & $0.270$ & $0.88$ & $0.327$ \\
                          & C\_mix & $0.678$ & $0.292$ & $0.58$ & $0.355$ \\
\midrule
\multirow{2}{*}{DFDC}     & B\_mix & $0.727$ & $0.368$ & $0.38$ & $0.455$ \\
                          & C\_mix & $0.728$ & $0.368$ & $0.30$ & $0.456$ \\
\midrule
\multirow{2}{*}{DFF}      & B\_mix & $0.551$ & $0.091$ & $0.90$ & $0.103$ \\
                          & C\_mix & $0.531$ & $0.060$ & $0.99$ & $0.061$ \\
\bottomrule
\end{tabular}
\end{table}

\subsection{Robustness: one condition where T-SBI helps}
Restricting attention to clean-image AUC would understate T-SBI. Under the image-level
perturbation grid (JPEG $q\in\{95,75,55,40\}$, Gaussian blur $\sigma\in\{0.5,1,2,3\}$,
denoise/sharpen, gamma, resize), the mixed T-SBI variant is the more compression-robust
model, and the effect is largest exactly where compression is heaviest
(Table~\ref{tab:robust}). On DFDC at $q40$, C\_mix retains frame AUC $0.780$ against
B\_mix's $0.696$---a margin of $+0.084$, the largest single-condition gap in the study
and an order of magnitude larger than any clean-image difference. C\_mix also degrades
more gracefully under heavy blur on all three face-swap corpora ($+0.02$--$0.03$ at
$\sigma=3$).

This is consistent with the score-distribution account rather than in tension with it.
Illumination-varied training pairs push the model toward low-frequency global cues,
which are precisely the cues that survive aggressive quantisation, whereas
blending-boundary evidence is high-frequency and is destroyed first. T-SBI therefore
changes \emph{which} evidence the model weights, which shows up as compression
robustness rather than as clean-image accuracy or as illumination-specific robustness.
We note that these runs share the single-seed limitation of the headline ablation.

\begin{table}[t]
\caption{Frame-level AUC under JPEG compression and Gaussian blur. C\_mix (T-SBI) is
markedly more robust to heavy compression on DFDC and degrades more gracefully under
blur on the face-swap corpora. Single-seed.}
\label{tab:robust}
\centering
\footnotesize
\setlength{\tabcolsep}{4pt}
\begin{tabular}{llrrrr}
\toprule
\textbf{Dataset} & \textbf{Model} & \textbf{$q95$} & \textbf{$q40$} & \textbf{$\sigma{=}0.5$} & \textbf{$\sigma{=}3$} \\
\midrule
\multirow{3}{*}{FF++}     & A      & $0.991$ & $0.984$ & $0.990$ & $\mathbf{0.925}$ \\
                          & B\_mix & $\mathbf{0.993}$ & $0.977$ & $\mathbf{0.993}$ & $0.843$ \\
                          & C\_mix & $0.990$ & $\mathbf{0.980}$ & $0.990$ & $0.875$ \\
\midrule
\multirow{3}{*}{Celeb-DF} & A      & $\mathbf{0.824}$ & $\mathbf{0.793}$ & $\mathbf{0.822}$ & $\mathbf{0.705}$ \\
                          & B\_mix & $0.747$ & $0.757$ & $0.745$ & $0.608$ \\
                          & C\_mix & $0.758$ & $0.760$ & $0.762$ & $0.627$ \\
\midrule
\multirow{3}{*}{DFDC}     & A      & $0.752$ & $0.730$ & $0.748$ & $\mathbf{0.610}$ \\
                          & B\_mix & $0.785$ & $0.696$ & $0.796$ & $0.575$ \\
                          & C\_mix & $\mathbf{0.793}$ & $\mathbf{0.780}$ & $\mathbf{0.800}$ & $0.606$ \\
\bottomrule
\end{tabular}
\end{table}

\subsection{The targeted attribute shows no effect}
\begin{figure}[t]
\centering
\includegraphics[width=\columnwidth]{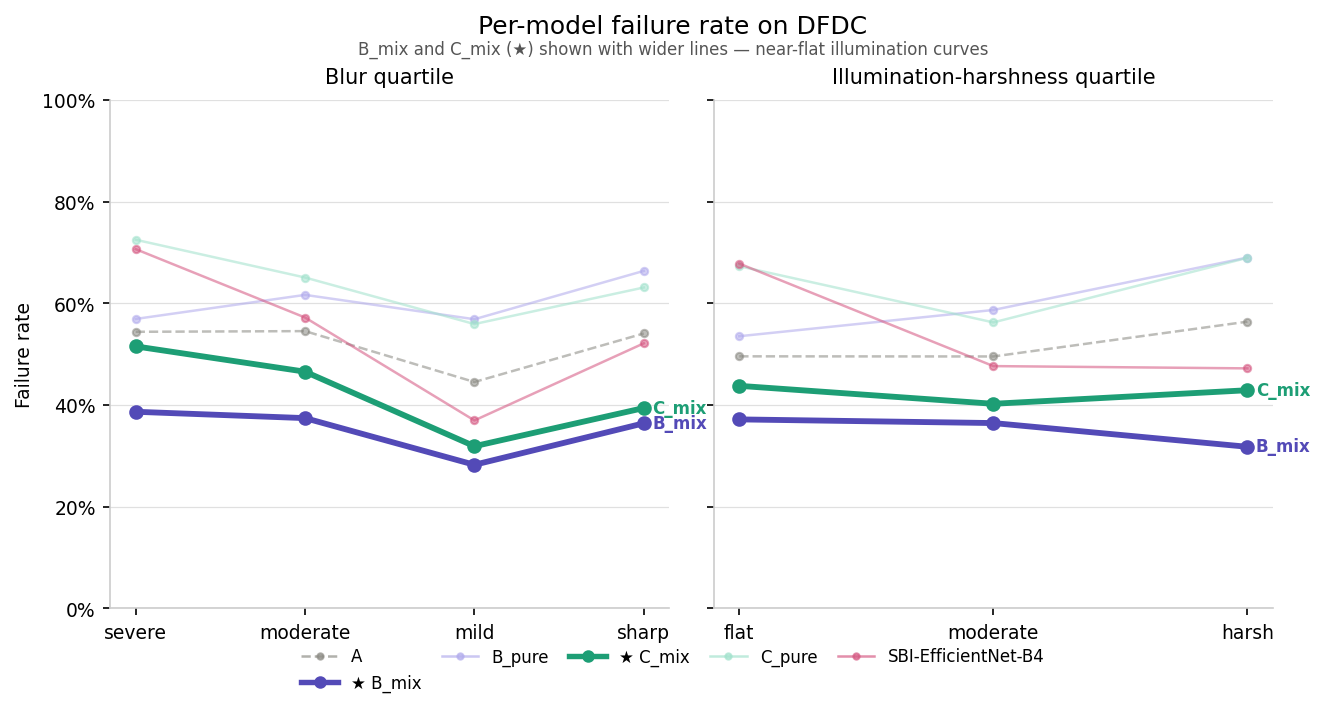}
\caption{Per-model failure rate on DFDC by blur quartile (left) and
illumination-harshness bin (right). B\_mix and C\_mix---the T-SBI contrast---are
near-flat on the illumination axis, with no preferential reduction on the harsh bin.}
\label{fig:tax}
\end{figure}

If T-SBI worked through illumination robustness, C\_mix should show either stronger
illumination structure in its failures than B\_mix, or a reduced failure rate on
harsh-illumination bins. Neither occurs. Across the taxonomy, C\_mix's
illumination-harshness Cram\'er's $V$ is \emph{lower} than B\_mix's on every dataset
(Celeb-DF $0.035$ vs $0.060$; DFDC $0.031$ vs $0.050$; FF++ $0.017$ vs $0.019$), all
below $0.05$---negligible---with per-bin failure-rate spreads of only $1$--$5$
percentage points and no monotonic ordering (Fig.~\ref{fig:tax}).

The wider taxonomy corroborates this. Of $216$ (model, dataset, axis) triples, only $23$
reach $V\ge0.10$, just $2$ exceed $0.20$, and none exceed $0.30$: there are no medium or
large effects anywhere. The strongest illumination-related sensitivity belongs not to
any T-SBI variant but to the off-the-shelf SBI reference on DFDC ($V=0.192$), and it
peaks on the \emph{flat}-illumination bin---the opposite of the T-SBI hypothesis. The
only clean monotonic effect in the predicted direction on any axis is head pose on FF++
(frontal $3.4$--$3.8\%$ vs profile $15.2$--$16.7\%$).

Failure-set overlap agrees: the B\_mix$\leftrightarrow$C\_mix Jaccard is $0.29$ on FF++
and $0.46$--$0.64$ on OOD corpora, so T-SBI changes \emph{which} samples fail at nearly
unchanged failure counts.

\section{Discussion}
Our three questions resolve negatively. The C-vs-B contrast changes sign across corpora;
an eightfold difference in training-pair $\Delta L$ produces frame-level differences
within seed noise on all four corpora; and the taxonomy shows no preferential
harsh-illumination improvement, with the T-SBI variant's illumination effect sizes
uniformly \emph{smaller} than its counterpart's.

What T-SBI reliably does is shift the score distribution, lowering the optimal decision
threshold by ${\approx}0.30$ where its effect is largest, and---relatedly---reweight the
evidence toward low-frequency cues, which is visible as a substantial
compression-robustness gain on DFDC. Both are real effects; neither is the illumination
mechanism the prior was designed around. This is not nothing: it changes
the false-positive/false-negative balance in a direction that favours fake-recall, which
some deployments want. But it is a calibration-level effect, and it would have been
easy to mistake for a capability gain---or, read at threshold $0.5$, for a capability
loss. Both readings appear in our own data, and only the threshold sweep distinguishes
them.

The generalizable lesson concerns validation practice. A prior motivated by a physical
attribute should be validated against that attribute, not against aggregate accuracy. An
attribute-binned failure taxonomy costs one inference pass over data already collected
and would have falsified the illumination mechanism immediately, whereas the aggregate
AUC deltas---$+0.014$ here, $-0.003$ there---invited over-interpretation in both
directions. The sampler diagnostic makes a related point: our gate admitted a median
pair below its own threshold while the corpus held far larger deltas, so a prior can be
much weaker in the realized training data than its design implies---reporting the
realized distribution of the gated quantity should be routine.

\textbf{Limitations.} The five-regime ablation is single-seed; only the
$\Delta L$-quartile diagnostic is repeated over three seeds, and small cross-regime AUC
deltas should be read with that in mind. The HIGHDL pool over-represents
DeepFakeDetection content while LOWDL over-represents FF++ YouTube footage, so pair
$\Delta L$ is partially confounded with source style. Our illumination descriptor
collapses spatial structure to two scalars and cannot distinguish side-lit from
front-lit faces at equal mean luminance---though it is the same descriptor the gate uses,
so the analysis is internally consistent. Finally, the taxonomy is computed at threshold
$0.5$; the operating-point analysis mitigates but does not remove this.

\section{Conclusion}
We built a temporal self-blending generator injecting a photometric prior into
face-swap detector training, then tried hard to detect its effect. Across a five-regime
ablation, a three-seed illumination-delta experiment, a dataset-level illumination
control, and a $506{,}328$-row failure taxonomy, the targeted mechanism is unsupported:
training-pair illumination delta does not measurably change frame-level detection, and no
preferential robustness appears on harsh-illumination bins. The effects that are present
are a ${\approx}0.30$ downward shift in the optimal decision threshold---which inverts the
apparent verdict depending on whether models are compared at $0.5$ or at their own
optima---and improved robustness to heavy compression, neither of which is the mechanism
the prior targeted.
We therefore report T-SBI as a controlled probe rather than an improvement, and offer its
diagnostics---realized-gate reporting, imbalance-robust threshold sweeps, and
attribute-binned failure taxonomies---as reusable instruments for evaluating any
physically motivated training prior.

\appendices
\section{Training and Evaluation Protocol}
\label{app:protocol}
Table~\ref{tab:protocol} consolidates the configuration shared by all five regimes and
both $\Delta L$-quartile variants, for reproducibility.

\begin{table}[h]
\caption{T-SBI pipeline, training, and evaluation configuration. All regimes share the
backbone, schedule, and augmentation; regimes differ only in the composition of the fake
class.}
\label{tab:protocol}
\centering
\footnotesize
\renewcommand{\arraystretch}{1.1}
\begin{tabular}{@{}p{0.36\columnwidth} p{0.58\columnwidth}@{}}
\toprule
\textbf{Item} & \textbf{Setting} \\
\midrule
\multicolumn{2}{@{}l}{\textit{T-SBI generation}}\\
Landmarks        & MediaPipe FaceMesh (ellipse fallback) \\
Mask             & Feathered convex hull over target face \\
Transfer         & Luminance-only histogram matching \\
Pair rejection   & ${>}1.3\times$ scale or ${>}35\%$ centre offset \\
Gate             & $\Delta L_{\text{mean}}\!\ge\!6$ or $\Delta L_{\text{std}}\!\ge\!3$, relax fallback \\
Accepted pairs   & $7{,}320$ from $732$ source videos \\
\midrule
\multicolumn{2}{@{}l}{\textit{Training}}\\
Backbone         & EfficientNet-B4 \\
Optimizer        & AdamW, cosine warmup \\
Input resolution & $380\times380$ \\
Epochs / batch   & $20$ / $32$ \\
Sampling         & Class-balanced within batch \\
JPEG randomization & $q\in[75,98]$, all classes, all regimes \\
Seeds            & $42$ (headline); $42/123/235$ ($\Delta L$ quartile) \\
\midrule
\multicolumn{2}{@{}l}{\textit{Evaluation}}\\
Corpora          & FF++ (in-dist.); Celeb-DF, DFDC, DFF (OOD) \\
Metrics          & AUC, ECE, balanced acc., MCC, Youden's J \\
Threshold sweep  & $99$ points in $[0.01,0.99]$ \\
Aggregation      & Frame; video max / mean / vote \\
Taxonomy         & $9$ attributes, within-dataset quantile bins \\
\bottomrule
\end{tabular}
\end{table}


\end{document}